\documentclass[11pt,a4paper]{article}
\usepackage[hyperref]{naaclhlt2019}

\usepackage{pgf}
\usepackage{tikz}
\usetikzlibrary{fit,positioning,arrows.meta,calc,matrix,chains,positioning,decorations.pathreplacing,decorations.markings}
\usepackage{tikz-dependency}
\usepackage{subcaption}
\usepackage{times}
\usepackage{latexsym}
\usepackage{amsmath}
\usepackage{multirow}
\usepackage{url}

\usepackage{amsfonts}
\usepackage{rotating}

\newcommand{\allproj}{{\cal P}}

\newtheorem{definition}[equation]{Definition}
\usepackage{color,colortbl}

\usepackage{xparse}
\newcounter{listtotal}\newcounter{listcntr}%
\usepackage{comment,mathtools,todonotes}

\aclfinalcopy 
\def\aclpaperid{377} 

\title{Low-Resource Syntactic Transfer with Unsupervised Source Reordering}

\author{Mohammad Sadegh Rasooli \\
  Facebook AI \\
  Menlo Park, CA, USA \\
  {\tt rasooli@fb.com} \\\And
  Michael Collins \\
 Department of Computer Science \\
  Columbia University \\
  {\tt mcollins@cs.columbia.edu} \\}

\date{}

\begin{document}

\maketitle

\begin{abstract}
    We describe a cross-lingual transfer method for dependency parsing that takes into account the problem of word order differences between source and target languages. Our model only relies on the Bible, a considerably smaller parallel data than the commonly used parallel data in transfer methods. We use the concatenation of projected trees from the Bible corpus, and the gold-standard treebanks in multiple source languages along with cross-lingual word representations. We demonstrate that reordering the source treebanks before training on them for a target language improves the accuracy of languages outside the European language family. Our experiments on 68 treebanks (38 languages) in the Universal Dependencies corpus achieve a high accuracy for all languages. Among them, our experiments on 16 treebanks of 12 non-European languages achieve an average UAS absolute improvement of $3.3\%$ over a  state-of-the-art method. 
\end{abstract}

 \section{Introduction}
There has recently been a great deal of interest in cross-lingual
transfer of dependency parsers, for which a parser is trained for a
target language of interest using treebanks in other languages.
Cross-lingual transfer can eliminate the need for the
expensive and time-consuming task of treebank annotation for
low-resource languages. Approaches include
annotation projection using parallel data
sets~\cite{hwa2005bootstrapping,ganchev-gillenwater-taskar:2009:ACLIJCNLP},
direct model transfer through learning of a delexicalized model from other
treebanks~\cite{zeman2008cross,tackstrom2013target}, treebank
translation~\cite{tiedemann-agic-nivre:2014:W14-16}, using synthetic
treebanks~\cite{tiedemann2016synthetic,wang2016galactic}, using
cross-lingual word
representations~\cite{tackstrom2012cross,guo2016representation,rasooli2017cross}
and using cross-lingual
dictionaries~\cite{durrett-pauls-klein:2012:EMNLP-CoNLL}.

Recent results from \newcite{rasooli2017cross} have shown accuracies
exceeding 80\% on unlabeled attachment accuracy (UAS) for several European
languages.\footnote{Specifically, Table 9 of \newcite{rasooli2017cross} shows
13 datasets, and 11 languages, with UAS scores of over 80\%; all of these
datasets are in European languages.}
However non-European languages remain a significant challenge for 
cross-lingual transfer. 
One hypothesis, which we investigate in this paper, is that word-order
differences between languages are a significant challenge for
cross-lingual transfer methods.
The main goal of our work is therefore to reorder gold-standard
source treebanks to make those treebanks syntactically more similar to
the target language of interest. We use two different approaches for
source treebank reordering: 1) reordering based on dominant dependency
directions according to the projected dependencies, 2) learning a
classifier on the alignment data. We show that an ensemble of these
methods with the baseline method leads to higher performance for the
majority of datasets in our experiments. We show particularly
significant improvements for non-European
languages.\footnote{Specifically, performance of our method gives an
  improvement of at least 2.3\% absolute scores in UAS on 11 datasets
  in 9 languages---Coptic, Basque, Chinese, Vietnamese, Turkish,
  Persian, Arabic, Indonesian Hebrew---with an average improvement of
  over 4.5\% UAS.}

The main contributions of this work are as follows:

\begin{itemize}
    \item We propose two different syntactic reordering methods based on the dependencies projected using translation alignments. The first model is based on the dominant dependency direction in the target language according to the projected dependencies. The second model learns a reordering classifier from the small set of aligned sentences in the Bible parallel data. 
    \item We run an extensive set of experiments on 68 treebanks for 38 languages. We show that by just using the Bible data, we are able to achieve significant improvements in non-European languages. Our ensemble method is able to  maintain a high accuracy in European languages. 
    \item We show that syntactic transfer methods can outperform a supervised model for cases in which the gold-standard treebank is very small. This indicates the strength of these models when the language is truly low-resource.
\end{itemize}

Unlike most previous work for which a simple delexicalized model with gold part-of-speech tags are used, we use lexical features and automatic part-of-speech tags. Our final model improves over two strong baselines, one with annotation projection and the other one inspired by the non-neural state-of-the-art model of \newcite{rasooli2017cross}. Our final results improve the performance on non-European languages by an average UAS absolute improvement of $3.3\%$ and LAS absolute improvement of $2.4\%$. 
 
 \section{Related Work}

There has recently been a great deal of research on dependency parser transfer.  Early work on direct model transfer~\cite{zeman2008cross,mcdonald-petrov-hall:2011:EMNLP,cohen-das-smith:2011:EMNLP,rosa-zabokrtsky:2015:ACL-IJCNLP,wang2018surface} considered learning a delexicalized parser from one or many source treebanks. A number of papers~\cite{naseem2012selective,tackstrom2013target,yuanregina15,ammar_tacl16,wang_topo} have considered making use of topological features to overcome the problem of syntactic differences across languages.  Our work instead reorders the source treebanks to make them similar to the target language before training on the source treebanks. 

\newcite{agic_selection} use  part-of-speech sequence similarity between the source and target language for selecting the source sentences in a direct transfer approach. \newcite{isomorphic_transfer} preprocess source trees to increase the isomorphy between the source and the target language dependency trees. They apply their method on a simple delexicalized model and their accuracy on the small set of languages that they have tried is significantly worse than ours in all languages. The recent work by \newcite{wang_emnlp18} reorders delexicalized treebanks of part-of-speech sequences in order to make it more similar to the target language of interest. The latter work is similar to our work in terms of using reordering. Our work is more sophisticated by using a full-fledged parsing model with automatic part-of-speech tags and every accessible dataset such as projected trees and multiple source treebanks as well as cross-lingual word embeddings for all languages.  

Previous work~\cite{tackstrom2012cross,duong-EtAl:2015:CoNLL,guo-EtAl:2015:ACL-IJCNLP2,guo2016representation,ammar_tacl16} has considered using cross-lingual word representations. A number of authors~\cite{durrett-pauls-klein:2012:EMNLP-CoNLL,rasooli2017cross} have used cross-lingual dictionaries. We also make use of cross-lingual word representations and dictionaries in this paper. We use the automatically extracted dictionaries from the Bible to translate words in the source treebanks to the target language. One other line of research in the delexicalized transfer approach is creating a synthetic treebank ~\cite{tiedemann2016synthetic,wang2016galactic,wang_emnlp18}.

Annotation projection~\cite{hwa2005bootstrapping,ganchev-gillenwater-taskar:2009:ACLIJCNLP,mcdonald-petrov-hall:2011:EMNLP,ma-xia:2014:P14-1,rasooli-collins:2015:EMNLP,lacroix-EtAl:2016:N16-1,agic2016multilingual} is another approach in parser transfer. In this approach, supervised dependencies are projected through word alignments and then used as training data. Similar to previous work~\cite{rasooli2017cross}, we make use of a combination of projected dependencies from annotation projection in addition to partially translated source treebanks. One other approach is treebank translation~\cite{tiedemann-agic-nivre:2014:W14-16} for which a statistical machine translation system is used to translate source treebanks to the target language. These models need a large amount of parallel data for having an accurate translation system. 

Using the Bible data goes back to the work of \newcite{diab2000statistical} and \newcite{Yarowsky:2001:IMT:1072133.1072187}. Recently there has been more interest in using the Bible data for different tasks, due to its availability for many languages~\cite{christodouloupoulos2014massively,agic2015if,agic2016multilingual,rasooli2017cross}. Previous work~\cite{ostling2017neural} has shown that the size of the Bible dataset does not provide a reliable machine translation model. Previous work in the context of machine translation~\cite{bisazza2016survey,daiber2016universal} presumes the availability of a parallel data that is often much larger than the Bible data.

\section{Baseline Model}\label{sec_background}

Our model trains on the concatenation of projected dependencies $\allproj$  and all of the source treebanks $\mathcal{T}_1 \ldots \mathcal{T}_k$. The projected data is from the set of projected dependencies for which at least $80\%$ of words have projected dependencies or there is a span of length $l\geq 5$ such that all words in that span achieve a projected dependency. This is the same as the definition of dense structures $\allproj_{80} \cup \allproj_{\geq 5}$ by \newcite{rasooli-collins:2015:EMNLP}. 

We use our reimplementation of the state-of-the-art neural biaffine graph-based parser of \newcite{dozat2016deep}\footnote{\url{https://github.com/rasoolims/universal-parser}}. Because many words in the projected dependencies do not have a head assignment, the parser ignores words without heads during training. Inspired by \newcite{rasooli2017cross}, we replace every word in the source treebanks with its most frequent aligned translation word from the Bible data in the target language. If that word does not appear in the Bible, we use the original word. That way, we have a code-switched data for which some of the words are being translated. In addition to fine-tuning the word embeddings, we use the fixed pre-trained cross-lingual word embeddings using the training approach of \newcite{rasooli2017cross} using the Wikipedia data and the Bible dictionaries.

 \section{Approach}



\begin{figure}[t!]
    \centering
    \begin{subfigure}{0.5\textwidth}
        \centering
    \input{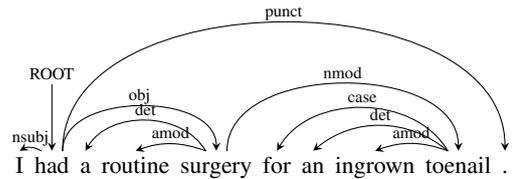}
    \caption{Original tree.}\label{fig:reordering_example_a}
    \end{subfigure}
      \begin{subfigure}{0.5\textwidth}
          \input{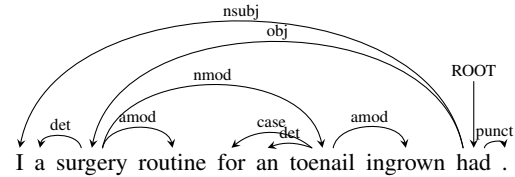}
    \caption{Persian-specific reordered tree.}\label{fig:reordering_example_b}
       \end{subfigure}
    \caption{An example of a gold-standard English tree that is reordered to look similar to the Persian syntactic order.}
    \label{fig:reordering_example}
\end{figure}
Before making use of the  source treebanks $\mathcal{T}_1 \ldots \mathcal{T}_k$ in the training data, we reorder each tree in the source treebanks to be syntactically more similar to the word order of the target language. In general, for a head $h$ that has $c$ modifiers $m_1 \ldots m_c$, we decide to put each of the dependents $m_i$ on the left or right of the head $h$. After placing them in the correct side of the head, the order in the original source sentence is preserved. Figure~\ref{fig:reordering_example} shows a real example of an English tree that is reordered for the sake of Persian as the target language. Here we see that we have a verb-final sentence, with nominal modifiers following the head noun. If one aims to translate this English sentence word by word, the reordered sentence gives a very good translation without any change in the sentence.

As mentioned earlier, we use two different approaches for source treebank reordering: 1) reordering based on dominant dependency directions according to the projected dependencies, 2) learning a classifier on the alignment data. We next describe these two methods.


\subsection{Model 1: Reordering Based on Dominant Dependency Direction}\label{sec_dominant}
The main goal of this model is to reorder source dependencies based on dominant dependency directions in the target language. We extract dominant dependency directions according to the projected dependencies $\allproj$ from the alignment data, and use the information for reordering source treebanks.

Let the tuple $\langle i, m, h, r\rangle$ show the dependency of the $m$'th word in the $i$'th projected sentence for which the $h$'th word is the parent with the dependency label $r$. $\langle i, m, {\tt NULL}, {\tt NULL}\rangle$ shows an unknown dependency for the $m$'th word: this occurs when some of the words in the target sentence do not achieve a projected dependency. We use the notations $h(i, m)$ and $r(i, m)$ to show the head index and dependency label of the $m$'th word in the $i$'th sentence.

\begin{definition} { Dependency direction:}
$d(i, m)$ shows the dependency direction of the $m$'th modifier word in the $i$'th sentence:

\[
d(i, m) = 
\begin{cases}
1 & {\tt if} ~~ h(i,m) > m \\
-1 & {\tt otherwise} \\
\end{cases}
\]
\end{definition}


\begin{definition}{ Dependency direction proportion: } Dependency direction proportion of each dependency label $l$ with direction $d \in \{-1, 1\}$ is defined as:

\[
\begin{split}
   \alpha^{(\allproj)} &(l, d) =  \\
   & \frac{\sum_{i=1}^{|\allproj|} \sum_{m=1}^{|\allproj^{(i)}|} {\mathbb{I}(r(i, m) = l ~\&~ d(i, m) = d)  } }{\sum_{i=1}^{|\allproj|} \sum_{m=1}^{|\allproj^{(i)}|} {\mathbb{I}(r(i, m) = l) }} 
\end{split}
\]

\end{definition}

\begin{definition}{  Dominant dependency direction:} 
For each dependency label $l$, we define the dominant dependency direction    $\lambda^{(\allproj)}(l) = d$ if $\alpha^{(\allproj)} (l, d) > 0.75$. In cases where there is no dominant dependency direction, $\lambda^{(\allproj)}(l) = 0$.

\end{definition}

We consider the following dependency labels for extracting  dominant dependency direction information: nsubj, obj, iobj, csubj, ccomp, xcomp, obl, vocative, expl, dislocated, advcl, advmod, aux, cop, nmod, appos, nummod, acl, amod. We find the direction of other dependency relations, such as  most of the function word dependencies and other non-core dependencies such as conjunction, not following a fixed pattern in the Universal Dependencies corpus.


\paragraph{Reordering condition}
Given a set of projections $\allproj$, we  calculate the dominant dependency direction information for the projections $\lambda^{(\allproj)}$. Similar to the projected dependencies, we  extract \emph{supervised} dominant dependency directions from the gold-standard source treebank $\mathcal{D}$: $\lambda^{(\mathcal{D})}$.  When we encounter a gold-standard dependency relation $\langle i, m, h, r\rangle$ in a source treebank $\mathcal{D}$, we change the direction if the following condition holds:
\[
\lambda^{(\mathcal{D})}(r) \neq \lambda^{(\allproj)}(r)  ~~{\tt and}~~ \lambda^{(\allproj)}(r) = -d(i,m)
\]

In other words, if the source and target languages do not have the same dominant dependency direction for $r$ and the dominant direction of the target language is the reverse of the current direction, we change the direction of that dependency. Reordering multiple dependencies in a gold standard tree then results in a reordering of the full tree, as for example in the transformation from Figure~\ref{fig:reordering_example_a} to Figure~\ref{fig:reordering_example_b}.



\subsection{Model 2: Reordering Classifier}\label{sec_reordering_classifier}
We now describe our approach for learning a reordering classifier for a target language using the alignment data. Unlike the first model for which we learn concrete rules, this model learns a reordering classifier from automatically aligned data. This model has two steps; the first step prepares the  training data from the automatically aligned parallel data, and the second step learns a classifier from the training data.

\subsubsection{Preparing Training Data from Alignments}
The goal of this step is to create training data for the reordering classifier. This data is extracted from the concatenation of parallel data from all source languages translated to the target language. Given a parallel dataset $(e^{(i)}, f^{(i)})$ for $i=1\ldots n$ that contains pairs of source and target sentences $e^{(i)}$ and $f^{(i)}$, the following steps are applied to create training data:
\begin{enumerate}
\item {\bf Extracting reordering mappings from alignments:} We first extract intersected word alignments for each source-target sentence pair. This is done by running the Giza++ alignments~\cite{och2000giza} in both directions. We ignore sentence pairs that more than half of the source words do not get alignment. We create a new mapping  $\mu^{(i)} = \mu^{(i)}_1\ldots \mu^{(i)}_{s_i}$ that maps each index $1 \leq j \leq s_i$ in the original source sentence to a unique index $1 \leq \mu^{(i)}_j \leq s_i$ in the reordered sentence. 
    \item {\bf Parsing source sentences:} We parse each source sentence using the supervised parser of the source language. We use the mapping $\mu^{(i)}$ to come up with a reordered tree for each sentence. In cases for which the number of non-projective arcs in the projected tree increase compared to the original tree, we do not use the sentence in the final training data. 
    \item {\bf Extracting classifier instances:} We create a training instance for every modifier word  $\langle i, m, h, r\rangle$. The decision about the direction of each dependency can be made based on the following condition:
\[
d^{*}(i, m) = 
\begin{cases}
1 & {\tt if} ~~\mu^{(i)}_{h}  > \mu^{(i)}_m \\
-1 & {\tt otherwise} \\
\end{cases}
\]
In other words, we decide about the new order of a dependency according to the mapping $\mu^{(i)}$. 
\end{enumerate}

\begin{figure}[t!]
    \centering
    
    \includegraphics[width=0.5\textwidth]{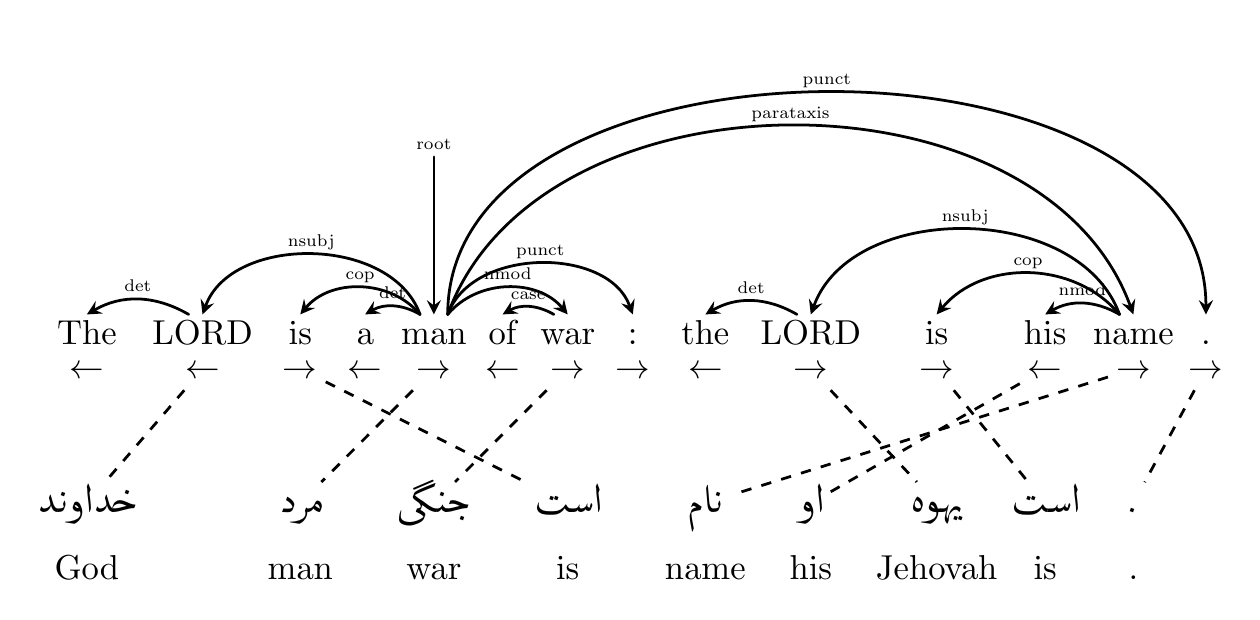}
    \caption{A reordering example from the Bible for English-Persian language pair. The Persian words are written from left to right for the ease of presentation. The arrows below the English words show the new dependency direction with respect to the word alignments to the Persian side. The reordered sentence would be ``The LORD a man of war is : his name the LORD is .''.}
    \label{fig:reordering_alignment}
\end{figure}

Figure~\ref{fig:reordering_alignment} shows an example for the data preparation step. As shown in the figure, the new directions for the English words are decided according to the Persian alignments.

\subsubsection{Classifier}
The reordering classifier  decides about the new direction of each dependency according to the recurrent representation of the head and dependent words. For a source sentence $e^{(i)} = e^{(i)}_1 \ldots e^{(i)}_{s_i}$ that belongs to a source language $\mathcal{L}$, we first obtain its recurrent representation $\eta^{(i)} = \eta^{(i)}_1 \ldots \eta^{(i)}_{s_i}$ by running a deep (3 layers) bi-directional LSTM~\cite{hochreiter1997long}, where $ \eta^{(i)}_j \in \mathbb{R}^{d_h}$.  For every dependency tuple $\langle i, m, h, r\rangle$, we use a multi-layer Perceptron (MLP) to decide about the new order $dir \in \{-1, 1\}$ of the $m$'th word with respect to its head $h$:
\[
p({\tt dir} | i, m, h, r) =  {\tt softmax_{dir}} (W \phi(i, m, h, r))
\]
where $W \in \mathbb{R}^{2 \times d_\phi}$ and $\phi(i, m, h, r) \in \mathbb{R}^{d_\phi}$ is as follows:
\[
\phi(i, m, h, r) = {\tt relu}( H q(i, m, h, r) + B )
\]
where {\tt relu} is the rectified linear unit activation~\cite{nair2010rectified}, $H \in \mathbb{R}^{d_\phi \times d_q}$, $B \in \mathbb{R}^{d_\phi}$, and $q(i, m, h, r) \in \mathbb{R}^{d_q}$ is as follows:
\[
q(i, m, h, r) = [\eta^{(i)}_m; \eta^{(i)}_h; R[r]; \Lambda[\mathbb{I}(h>m)]; L[\mathcal{L}]]
\]
where $\eta^{(i)}_m$ and $\eta^{(i)}_h$ are the recurrent representations for the modifier and head words respectively, $R$ is the dependency relation embedding dictionary that embeds every dependency relation to a $\mathbb{R}^{d_r}$ vector, $\Lambda$ is the direction embedding for the original position of the head with respect to its head and embeds each direction to a 2-dimensional vector, and $L$ is the language embedding dictionary that embeds the source language id $\mathcal{L}$ to a $\mathbb{R}^{d_L}$ vector.

The input to the recurrent layer is the concatenation of two input vectors. The first vector is the sum of the fixed pre-trained cross-lingual embeddings, and randomly initialized word vector. The second vector is the  part-of-speech tag embeddings. 

 Figure~\ref{fig:reordering_models} shows a graphical depiction of the two reordering models that we use in this work.

\NewDocumentCommand{\names}{o}{%
	\setcounter{listtotal}{0}\setcounter{listcntr}{-1}%
	\renewcommand*{\do}[1]{\stepcounter{listtotal}}%
	\expandafter\docsvlist\expandafter{\namesarray}%
	\IfNoValueTF{#1}
	{\namesarray}
	{
		\renewcommand*{\do}[1]{\stepcounter{listcntr}\ifnum\value{listcntr}=#1\relax##1\fi}%
		\expandafter\docsvlist\expandafter{\namesarray}}%
}

\NewDocumentCommand{\denames}{o}{%
	\setcounter{listtotal}{0}\setcounter{listcntr}{-1}%
	\renewcommand*{\do}[1]{\stepcounter{listtotal}}%
	\expandafter\docsvlist\expandafter{\denamesarray}%
	\IfNoValueTF{#1}
	{\namesarray}
	{
		\renewcommand*{\do}[1]{\stepcounter{listcntr}\ifnum\value{listcntr}=#1\relax##1\fi}%
		\expandafter\docsvlist\expandafter{\denamesarray}}%
}

\tikzset{neuron/.style={shape=circle, minimum size=.5cm, 
		inner sep=0, draw, font=\small}, io/.style={neuron, fill=gray!20}}
\tikzset{
	redondo/.style={
		draw=black,
		line width=1pt,
		rounded corners=3pt,
		text width=#1
	},
	punto/.style={
		fill=black,
		circle,
		inner sep=1.25pt
	},
	tresp/.pic={
		\node[punto] at (0.25,0) {};
		\node[punto] at (0.5,0) {};
		\node[punto] at (0.75,0) {};
	},
	dosp/.pic={
		\node[punto] at (0.3,0) {};
		\node[punto] at (0.4,0) {};
		\node[punto] at (0.5,0) {};
	},
	one_cir/.pic={
		\node[punto] at (0.25,0) {};
	},
	cuadra/.style={
		fill=teal,
		minimum size=10pt
	},
	arr/.style={
		line width=1pt,
		draw=green!70!black,
		->,
		>=latex
	}  
}

\tikzset{
	block filldraw/.style={
		draw},
	block rect/.style={
		block filldraw, rectangle},
	block/.style={
		block rect, minimum height=0.8cm, minimum width=6em},
	from/.style args={#1 to #2}{
		above right={0cm of #1},
		/utils/exec=\pgfpointdiff
		{\tikz@scan@one@point\pgfutil@firstofone(#1)\relax}
		{\tikz@scan@one@point\pgfutil@firstofone(#2)\relax},
		minimum width/.expanded=\the\pgf@x,
		minimum height/.expanded=\the\pgf@y}}

\begin{figure}[t!]
    \centering
   
   \scalebox{1.0}{
\begin{tikzpicture}[x=2cm, y=1.5cm, >=Stealth]

	\node [rectangle] at (0, 0) (x-1) {\tiny I};
	\node [rectangle] at (.4, 0) (x-2) {\tiny had};
	\node  [rectangle] at (0.8, 0)  (x-3) {\tiny a};
	\node[rectangle] at (1.2, 0) (x-4) {\tiny routine};
	\node[rectangle] at (1.6, 0) (x-5) {\tiny surgery};
	\node[rectangle] at (2.0, 0) (x-6) {\tiny for};
	\node[rectangle] at (2.4, 0) (x-7) {\tiny an};
\node[rectangle] at (2.8, 0) (x-8) {\tiny ingrown};
\node[rectangle] at (3.2, 0) (x-9) {\tiny nail};
\node[rectangle] at (3.6, 0) (x-10) {\tiny .};

	\foreach \n [evaluate={\nm=-0.4 + \n/2.5 ;}]  in {1, 2,3, 4,5,6,7,8,9,10}{
	\node [rectangle,draw,rounded corners=3pt,text=blue] at (\nm, .5)  (h-\n)  {\tiny $\eta_{\n}$}; 
		\draw [->] (x-\n.north) -- (h-\n.south);
}

	\foreach \n  [evaluate={\nm=int(\n-1);}] in {2,3, 4,5,6,7,8,9,10}{
		\draw [<->] (h-\n.west) -- (h-\nm.east);
}

  \node[rectangle,draw,rounded corners=3pt,text=blue] at (3.5, 1) (L) {\tiny L[en]};
  \node[rectangle,draw,rounded corners=3pt,text=blue] at (3.0, 1) (lambda) {\tiny $\lambda[-1]$};
  \node[rectangle,draw,rounded corners=3pt,text=blue] at (2.5, 1) (R) {\tiny R[obj]};

   \node[redondo=0.7cm,label={ left:{\tiny concat}}] at (2, 1.6) (cc)  {};
      \pic at ([xshift=-0.8cm]cc) {dosp};  	  
      
    \draw [->] (h-2.north) -- (cc.south);  
    \draw [->] (h-5.north) -- (cc.south);  
    \draw [->] (R.north) -- (cc.south);  
    \draw [->] (L.north) -- (cc.south);  
    \draw [->] (lambda.north) -- (cc.south);  
		  \node[rectangle,draw,rounded corners=3pt] at (2, 2.2) (H) {H};
\draw [->] (cc.north) -- (H.south)  node[midway,fill=white] {\tiny $\times$};	

		  \node[rectangle] at (2.2, 2.2) (plus) {+};
\node[rectangle,draw,rounded corners=3pt] at (2.4, 2.2) (B) {B};

		  \node[rectangle] at (2.2, 2.95) (W) {W};
\draw [->] (plus.north) -- (W.south)  node[midway,fill=white] {\tiny {\tt relu}};	

		  \node[rectangle] at (3.5, 2.95) (dir) {\tiny $dir = 1$};
\draw [->] (W.east) -- (dir.west)  node[midway,fill=white] {\tiny $\arg\max$};	

 \node[rectangle] at (1.5, -.5) (rel) {\tiny \tt obj};

    \draw [->] (x-2.south) -- (rel.west);  
    \draw [->] (x-5.south) -- (rel.east);

 \node[rectangle] at (1, -.9) (l_en) {\tiny $\lambda^{(\mathcal{D}^{en})}(obj) = -1$};

 \node[rectangle] at (2.5, -.9) (l_fa) {\tiny $\lambda^{(\mathcal{P}^{fa})}(obj) = 1$};
     \draw [->] (rel.south) -- (l_en.north);
     \draw [->] (rel.south) -- (l_fa.north);

 \node[rectangle] at (1.75, -1.4) (man_dir) {\tiny $dir = 1$};

    \draw [->] (l_en.south) -- (man_dir.north);
     \draw [->] (l_fa.south) -- (man_dir.north);
     
	\end{tikzpicture}
}

 \caption{Two different approaches for reordering the dependency order for the example in Figure~\ref{fig:reordering_example}. The reordering classifier is shown on top, for the dependency relation between the words ``had'' and ``surgery'' with an ``obj'' relation. At the bottom, the reordering model based on dominant dependency direction information is shown.}
    \label{fig:reordering_models}
\end{figure}
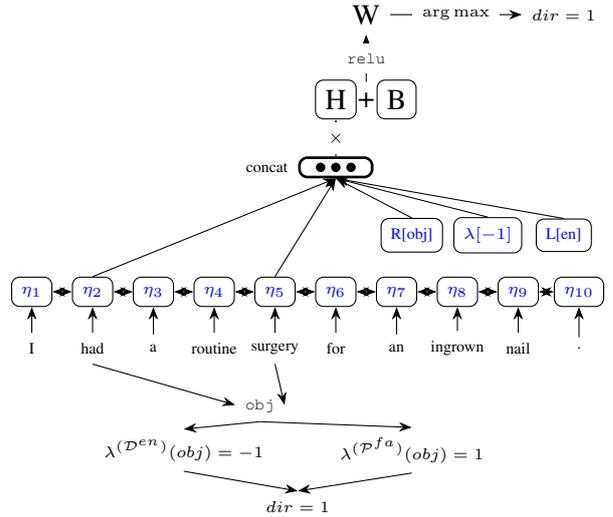

 \section{Experiments}

\paragraph{Datasets and Tools}
We use 68 datasets from 38 languages in the Universal Dependencies corpus version 2.0~\cite{universal_2}. The languages are Arabic (ar), Bulgarian (bg), Coptic (cop), Czech (cs), Danish (da), German (de), Greek (el), English (en), Spanish (es), Estonian (et), Basque (eu), Persian (fa), Finnish (fi), French (fr), Hebrew (he), Hindi (hi), Croatian (hr), Hungarian (hu), Indonesian (id), Italian (it), Japanese (ja), Korean (ko), Latin (la), Lithuanian (lt), Latvian (lv), Dutch (nl), Norwegian (no), Polish (pl), Portuguese (pt), Romanian (ro), Russian (ru), Slovak (sk), Slovene (sl), Swedish (sv), Turkish (tr), Ukrainian (uk), Vietnamese (vi), and Chinese (zh). 

We use the Bible data from \newcite{christodouloupoulos2014massively} for the 38 languages. We extract word alignments using Giza++ default model~\cite{och2000giza}. Following \newcite{rasooli-collins:2015:EMNLP}, we obtain intersected alignments and apply soft POS consistency to filter potentially incorrect alignments. We use the Wikipedia dump data to extract monolingual data for the languages in order to train monolingual embeddings. We follow the method of ~\newcite{rasooli2017cross} to use the extracted dictionaries from the Bible and monolingual text from Wikipedia to create cross-lingual word embeddings. We use the UDPipe pretrained models~\cite{udpipe:2017} to tokenize Wikipedia, and a reimplementation of the Perceptron tagger of \newcite{collins:2002:EMNLP02}\footnote{\url{https://github.com/rasoolims/SemiSupervisedPosTagger}} to achieve automatic POS tags trained on the training data of the Universal Dependencies corpus~\cite{universal_2}. We use word2vec~\cite{mikolov2013efficient}\footnote{\url{https://github.com/dav/word2vec}} to achieve embedding vectors both in monolingual and cross-lingual settings.

\paragraph{Supervised Parsing Models}
We trained our supervised models on the union of all datasets in a language to obtain a supervised model for each language. It is worth noting that there are two major changes that we make to the neural parser of \newcite{dozat2016deep} in our implementation\footnote{\url{https://github.com/rasoolims/universal-parser}} using the Dynet library~\cite{neubig2017dynet}: first, we add a one-layer character BiLSTM to represent the character information for each word. The final character representation is obtained by concatenating the forward representation of the last character and the backward representation of the first character. The concatenated vector is summed with the randomly initialized as well as fixed pre-trained cross-lingual word embedding vectors. Second, inspired by \newcite{weiss-EtAl:2015:ACL-IJCNLP}, we maintain the moving average parameters to obtain more robust parameters at decoding time. 

We excluded the following languages from the set of source languages for \emph{annotation projection} due to their low supervised accuracy: Estonian, Hungarian, Korean, Latin, Lithuanian, Latvian, Turkish, Ukrainian, Vietnamese, and Chinese.

\paragraph{Baseline Transfer Models}
We use two baseline models: 1)  Annotation projection: This model only trains on the projected dependencies. 2) Annotation projection + direct transfer: To speed up training, we sample at most thousand sentences from each treebank, comprising a training data of about 37K sentences. 

\subsection{Reordering Ensemble Model}
We noticed that our reordering models perform better in non-European languages, and perform slightly worse in European languages. We use the following ensemble model to make use of all of the three models (annotation projection + direct transfer, and the two reordering models), to make sure that we always obtain an accurate parser.  

The ensemble model is as follows: given three output trees for the $i$'th sentence $\langle i_j, m, h_j, r_j\rangle$ for $j=1,2,3$ in the target language $\mathcal{L}$, where the first tuple ($j=1$) belongs to the baseline model, the second ($j=2$) and third ($j=3$) belong to the two reordering models, we weight each dependency edge with respect to the following conditions:

\[
\omega(m, h, r)   = z(m, h, r) \cdot
     \sum_{j=1}^{3} c(j, \mathcal{L}) \cdot  \mathbb{I}(\langle i_j, m, h, r\rangle) 
\]

where $c(j, \mathcal{L})$ is a coefficient that puts more weight on the first or the other two outputs depending on the target language family:

\[
c(j, \mathcal{L}) = 
\begin{cases}
2 & \text{\tt if } j = 1 ~\&~  \mathcal{L} ~\text{\tt is European} \\
2 & \text{\tt if } j > 1 ~\&~  \mathcal{L} ~\text{\tt is not European}\\
1 & \text{\tt otherwise} \\
\end{cases}
\]

and $z(m, h, r)$ is a simple weighting depending on the dominant order information:

\[
z(m, h, r) = 
\begin{cases}
1 & \text{if } direction(\langle m, h\rangle) = -\lambda^{(\allproj)}(r) \\
3 & \text{if } direction(\langle m, h\rangle) = \lambda^{(\allproj)}(r)\\
2 & \text{otherwise}~ (\lambda^{(\allproj)}(r) = 0)\\
\end{cases}
\]

The above coefficients are modestly tuned on the Persian language as our development language. We have not seen any significant change in modifying the numbers: instead, the fact that an arc with a dominant dependency direction is regarded as a more valuable arc, and the baseline should have more effect in the European languages suffices for the ensemble model. 

We run the Eisner first-order graph-based algorithm~\cite{eisner1996three} on top of the edge weights $\omega$ to extract the best possible tree. 

\subsection{Parameters}
We run all of the transfer models with 4000 mini-batches, in which each mini-batch contains approximately 5000 tokens. We follow the same parameters as in \newcite{dozat2016deep} and use a dimension of 100 for character embeddings. For the reordering classifier, we use the Adam algorithm~\cite{adam_paper} with default parameters to optimize the log-likelihood objective. We filter the alignment data to keep only those sentences for which at least half of the source words have an alignment. We randomly choose $1\%$ of the reordering data as our heldout data for deciding when to stop training the reordering models. Table~\ref{tab:reorder_params} shows the parameter values that we use in the reordering classifier.

\begin{table}[t!]
    \centering
        \setlength{\tabcolsep}{3pt}

    \begin{tabular}{l|c | c}
    \hline \hline
         Variable & Notation & Size  \\ \hline
        Word embedding & $d_w$ & 100 \\
        POS embedding & $d_p$ & 100 \\
        Bi-LSTM & $d_h$ & 400 \\
        Dep. relation embedding & $d_r$ & 50 \\
        Language ID embedding & $d_L$ & 50 \\
        Hidden layer & $d_{\phi}$ & 200 \\ 
      
        Number of BiLSTM layers &  -- & 3 \\
        Mini-batch size (tokens) & -- & $\sim 1000$    \\
                \hline \hline
    \end{tabular}
    \caption{Parameter values in the reordering classifier model.}
    \label{tab:reorder_params}
\end{table}
\definecolor{LightCyan}{rgb}{0.88,1,1}

\begin{table*}[t!]
    \centering
    \footnotesize
     \renewcommand{\arraystretch}{0.75}
     \setlength{\tabcolsep}{4.5pt}
    \begin{tabular}{l | c c | c c || c c | c c |  c c | c c ||  c c   }
    \hline \hline
      \multirow{3}{*}{Dataset}   & \multicolumn{4}{c||}{Baselines} & \multicolumn{8}{c || }{Reordering} & \multicolumn{2}{c}{      \multirow{2}{*}{Supervised}} \\
      &  \multicolumn{2}{c}{Projection}   &  \multicolumn{2}{c||}{Direct+Proj }  &  \multicolumn{2}{c}{Dominant}  &  \multicolumn{2}{c}{Classifier}  & \multicolumn{2}{c}{ Ensemble}   &  \multicolumn{2}{c||}{\bf Difference}   & &    \\ \cline{2-15}
  &      UAS  &  LAS  &  UAS  &  LAS  &  UAS  &  LAS  &  UAS  &  LAS  &  UAS  &  LAS  & UAS & LAS &  UAS  &  LAS  \\   \hline
\rowcolor{LightCyan} Coptic & 2.0 & 0.4 & 58.5 & 37.6 & 69.1 & 52.7 & 65.5 & 50.9 & 69.6 & 52.7 & 11.1 & 15.1 & 86.9 & 80.1 \\ 
\rowcolor{LightCyan} Basque & 39.5 & 22.0 & 44.9 & 29.0 & 53.7 & 34.0 & 48.6 & 32.2 & 53.7 & 34.4 & 8.8 & 5.4 & 81.9 & 75.9 \\ 
\rowcolor{LightCyan} Chinese & 23.6 & 10.8 & 40.6 & 17.8 & 47.3 & 25.4 & 45.4 & 23.5 & 47.0 & 25.6 & 6.4 & 7.8 & 81.1 & 74.8 \\ 
\rowcolor{LightCyan} Vietnamese & 44.6 & 26.8 & 51.2 & 33.6 & 55.3 & 34.5 & 50.4 & 34.2 & 55.1 & 34.5 & 4.0 & 0.9 & 66.2 & 56.7 \\ 
\rowcolor{LightCyan} Turkish\_pud & 44.7 & 19.9 & 46.6 & 24.5 & 50.3 & 26.7 & 42.6 & 22.0 & 49.9 & 26.3 & 3.4 & 1.8 & 56.7 & 31.7 \\ 
\rowcolor{LightCyan} Persian & 54.4 & 46.2 & 61.8 & 53.0 & 64.3 & 54.7 & 63.0 & 53.4 & 65.1 & 55.4 & 3.3 & 2.4 & 87.8 & 83.6 \\ 
\rowcolor{LightCyan} Arabic\_pud & 60.3 & 44.2 & 65.2 & 50.5 & 68.2 & 52.0 & 66.5 & 51.4 & 68.3 & 52.3 & 3.2 & 1.8 & 71.9 & 58.8 \\ 
\rowcolor{LightCyan} Indonesian & 59.9 & 42.8 & 72.1 & 56.0 & 73.6 & 56.5 & 72.9 & 56.8 & 74.6 & 56.7 & 2.5 & 0.6 & 84.8 & 77.4 \\ 
\rowcolor{LightCyan} Turkish & 44.6 & 23.9 & 46.6 & 29.3 & 48.9 & 30.6 & 44.9 & 26.6 & 49.0 & 30.0 & 2.4 & 0.7 & 64.2 & 52.5 \\ 
\rowcolor{LightCyan} Hebrew & 63.1 & 46.9 & 70.4 & 55.4 & 72.4 & 54.9 & 71.6 & 55.7 & 72.7 & 55.4 & 2.3 & 0.0 & 88.2 & 82.4 \\ 
\rowcolor{LightCyan} Arabic & 49.5 & 36.8 & 58.9 & 46.8 & 60.8 & 48.3 & 59.2 & 46.9 & 61.2 & 48.8 & 2.3 & 2.0 & 85.6 & 78.9 \\ 
\rowcolor{LightCyan} Japanese & 54.8 & 38.9 & 65.2 & 46.5 & 65.9 & 46.8 & 64.1 & 44.8 & 66.6 & 46.8 & 1.4 & 0.3 & 94.5 & 92.7 \\ 
\rowcolor{LightCyan} Japanese\_pud & 58.6 & 44.1 & 66.8 & 51.5 & 67.4 & 51.5 & 64.7 & 48.4 & 67.9 & 51.9 & 1.1 & 0.4 & 94.7 & 93.5 \\ 
\rowcolor{LightCyan} Korean & 34.3 & 17.3 & 43.0 & 24.8 & 43.5 & 23.8 & 43.6 & 26.4 & 44.1 & 24.7 & 1.1 & -0.2 & 76.2 & 69.9 \\ 
\rowcolor{LightCyan} Hindi\_pud & 53.4 & 43.3 & 58.2 & 47.6 & 58.3 & 47.5 & 58.8 & 48.5 & 58.9 & 48.2 & 0.6 & 0.6 & 70.2 & 55.6 \\ 
\rowcolor{pink} Lithuanian & 60.6 & 42.5 & 66.6 & 49.5 & 63.7 & 46.8 & 64.6 & 46.0 & 67.2 & 49.9 & 0.6 & 0.4 & 54.8 & 40.0 \\ 
Czech\_cac & 33.9 & 14.8 & 76.2 & 66.9 & 76.3 & 66.7 & 75.2 & 65.8 & 76.7 & 67.4 & 0.5 & 0.6 & 92.1 & 88.3 \\ 
Czech\_cltt & 13.7 & 5.1 & 69.4 & 59.7 & 69.7 & 59.5 & 66.6 & 57.8 & 70.0 & 60.3 & 0.5 & 0.6 & 88.9 & 84.9 \\ 
French\_partut & 81.6 & 75.2 & 84.3 & 77.8 & 84.9 & 78.4 & 84.4 & 78.1 & 84.8 & 78.4 & 0.5 & 0.5 & 90.0 & 85.1 \\ 
Croatian & 70.6 & 59.9 & 79.4 & 69.9 & 79.3 & 69.5 & 77.9 & 67.7 & 79.9 & 70.1 & 0.5 & 0.2 & 86.8 & 80.4 \\ 
Greek & 62.3 & 47.2 & 75.9 & 63.9 & 75.4 & 63.1 & 74.7 & 62.5 & 76.4 & 64.1 & 0.4 & 0.2 & 88.0 & 84.4 \\ 
Russian\_pud & 75.7 & 65.8 & 81.1 & 72.2 & 80.9 & 72.2 & 79.9 & 70.7 & 81.5 & 72.7 & 0.4 & 0.5 & 86.5 & 74.1 \\ 
German & 71.4 & 62.3 & 75.4 & 67.1 & 75.6 & 67.1 & 75.5 & 66.4 & 75.8 & 67.3 & 0.4 & 0.2 & 85.9 & 81.2 \\ 
French & 80.2 & 72.9 & 83.0 & 75.9 & 82.9 & 75.9 & 83.3 & 75.9 & 83.4 & 76.2 & 0.4 & 0.3 & 90.4 & 86.9 \\ 
Czech & 33.9 & 14.5 & 74.6 & 65.3 & 74.1 & 64.4 & 73.0 & 63.7 & 75.0 & 65.8 & 0.4 & 0.5 & 92.5 & 89.1 \\ 
Finnish\_pud & 64.1 & 52.5 & 67.2 & 55.0 & 66.8 & 55.0 & 67.3 & 55.1 & 67.5 & 55.5 & 0.4 & 0.5 & 81.6 & 74.5 \\ 
Dutch & 59.2 & 48.2 & 68.5 & 55.2 & 69.6 & 55.9 & 68.3 & 54.4 & 68.8 & 55.4 & 0.4 & 0.1 & 83.5 & 76.6 \\ 
Russian & 68.9 & 59.4 & 75.1 & 63.9 & 75.4 & 64.1 & 74.5 & 63.4 & 75.5 & 64.3 & 0.4 & 0.4 & 85.7 & 77.9 \\ 
Latin\_ittb & 56.4 & 42.5 & 63.0 & 49.2 & 63.2 & 49.5 & 62.4 & 48.7 & 63.3 & 49.7 & 0.4 & 0.4 & 89.5 & 86.5 \\ 
Norwegian\_nynorsk & 72.5 & 62.9 & 76.4 & 68.1 & 76.5 & 68.0 & 76.1 & 67.3 & 76.8 & 68.4 & 0.3 & 0.3 & 91.3 & 88.8 \\ 
\rowcolor{pink} Ukrainian & 55.1 & 36.9 & 64.3 & 46.1 & 64.5 & 45.7 & 61.7 & 42.2 & 64.6 & 45.9 & 0.3 & -0.2 & 43.3 & 22.1 \\ 
Bulgarian & 80.4 & 69.4 & 83.8 & 73.8 & 84.0 & 73.8 & 83.1 & 73.0 & 84.1 & 73.9 & 0.3 & 0.1 & 90.9 & 86.0 \\ 
English\_lines & 75.6 & 66.5 & 77.8 & 69.0 & 78.9 & 69.9 & 77.0 & 68.2 & 78.1 & 69.2 & 0.3 & 0.3 & 85.8 & 80.5 \\ 
Finnish\_ftb & 63.9 & 46.5 & 66.0 & 48.3 & 65.8 & 47.6 & 65.7 & 48.1 & 66.3 & 48.4 & 0.3 & 0.1 & 81.1 & 74.4 \\ 
Russian\_syntagrus & 69.4 & 57.5 & 73.9 & 62.2 & 73.8 & 61.8 & 73.2 & 61.2 & 74.2 & 62.3 & 0.3 & 0.1 & 91.3 & 88.3 \\ 
Finnish & 60.6 & 48.7 & 64.6 & 51.9 & 63.5 & 51.2 & 63.7 & 51.1 & 64.8 & 52.0 & 0.2 & 0.1 & 80.9 & 73.5 \\ 
Hungarian & 58.3 & 41.1 & 67.8 & 49.0 & 67.8 & 48.9 & 65.8 & 47.4 & 68.0 & 49.1 & 0.2 & 0.1 & 78.2 & 69.8 \\ 
Czech\_pud & 35.7 & 16.6 & 77.5 & 69.3 & 76.7 & 67.6 & 76.2 & 67.7 & 77.7 & 69.4 & 0.2 & 0.2 & 89.9 & 84.4 \\ 
Dutch\_lassysmall & 61.8 & 52.1 & 73.9 & 63.4 & 73.8 & 62.8 & 73.0 & 61.9 & 74.0 & 63.3 & 0.2 & 0.0 & 91.3 & 87.3 \\ 
Slovenian\_sst & 58.4 & 44.1 & 61.7 & 47.7 & 61.6 & 47.7 & 61.6 & 47.4 & 61.9 & 48.0 & 0.2 & 0.3 & 70.6 & 63.6 \\ 
English\_pud & 73.5 & 65.5 & 75.9 & 69.3 & 77.1 & 69.9 & 74.5 & 67.7 & 76.0 & 69.4 & 0.2 & 0.2 & 88.3 & 84.2 \\ 
German\_pud & 74.1 & 65.3 & 77.8 & 68.9 & 77.7 & 68.5 & 76.9 & 67.4 & 78.0 & 68.8 & 0.1 & 0.0 & 85.9 & 79.0 \\ 
Polish & 77.6 & 64.7 & 79.9 & 67.9 & 79.7 & 67.5 & 79.5 & 67.2 & 80.1 & 68.0 & 0.1 & 0.1 & 89.4 & 83.3 \\ 
Swedish\_lines & 77.2 & 67.7 & 81.1 & 71.6 & 80.7 & 71.1 & 80.1 & 70.4 & 81.3 & 71.7 & 0.1 & 0.1 & 86.9 & 81.5 \\ 
English & 70.1 & 61.6 & 72.8 & 64.6 & 73.5 & 65.2 & 71.6 & 63.5 & 72.9 & 64.8 & 0.1 & 0.3 & 88.2 & 84.8 \\ 
Spanish & 78.5 & 68.0 & 83.1 & 73.8 & 83.2 & 73.8 & 82.3 & 72.8 & 83.2 & 73.9 & 0.1 & 0.1 & 89.3 & 83.9 \\ 
Swedish & 75.3 & 67.0 & 79.0 & 70.9 & 78.8 & 70.9 & 78.2 & 70.0 & 79.1 & 71.0 & 0.1 & 0.1 & 86.7 & 82.3 \\ 
English\_partut & 72.0 & 65.3 & 77.4 & 71.1 & 78.0 & 71.1 & 76.3 & 69.9 & 77.5 & 71.2 & 0.1 & 0.1 & 88.4 & 83.0 \\ 
Swedish\_pud & 75.9 & 67.4 & 80.5 & 72.1 & 80.2 & 72.0 & 79.2 & 71.0 & 80.6 & 72.1 & 0.1 & 0.0 & 84.0 & 77.6 \\ 
Italian & 81.3 & 74.4 & 85.0 & 79.0 & 85.4 & 79.5 & 84.4 & 78.1 & 85.1 & 79.1 & 0.1 & 0.0 & 92.1 & 89.5 \\ 
Romanian & 72.8 & 59.0 & 76.8 & 64.2 & 76.2 & 63.7 & 75.3 & 63.2 & 76.8 & 64.3 & 0.1 & 0.1 & 89.6 & 83.5 \\ 
Estonian & 63.1 & 40.8 & 66.7 & 46.0 & 65.6 & 45.8 & 65.5 & 45.2 & 66.7 & 46.1 & 0.1 & 0.2 & 71.6 & 60.7 \\ 
Portuguese & 62.6 & 50.7 & 84.1 & 76.9 & 83.7 & 76.6 & 83.4 & 76.2 & 84.2 & 77.1 & 0.0 & 0.2 & 90.6 & 85.6 \\ 
Portuguese\_br & 60.6 & 47.7 & 81.3 & 71.2 & 80.8 & 70.8 & 80.8 & 70.4 & 81.4 & 71.3 & 0.0 & 0.2 & 91.6 & 89.0 \\ 
Norwegian\_bokmaal & 78.0 & 70.5 & 80.5 & 73.2 & 80.6 & 73.4 & 79.7 & 72.1 & 80.5 & 73.2 & 0.0 & 0.0 & 92.1 & 89.7 \\ 
French\_pud & 81.0 & 72.8 & 83.7 & 75.7 & 84.2 & 76.2 & 83.3 & 75.2 & 83.7 & 75.7 & 0.0 & 0.0 & 89.1 & 83.8 \\ 
Spanish\_pud & 81.3 & 70.9 & 84.3 & 75.6 & 84.6 & 76.0 & 83.6 & 74.6 & 84.3 & 75.7 & 0.0 & 0.1 & 89.1 & 80.8 \\ 
Latvian & 59.0 & 43.6 & 63.3 & 47.2 & 62.1 & 45.6 & 60.7 & 44.7 & 63.3 & 47.0 & 0.0 & -0.2 & 71.3 & 61.2 \\ 
Italian\_pud & 83.8 & 76.0 & 87.3 & 81.3 & 87.5 & 81.3 & 86.5 & 79.9 & 87.3 & 81.2 & 0.0 & -0.1 & 91.9 & 88.4 \\ 
French\_sequoia & 79.1 & 73.0 & 82.2 & 76.4 & 81.6 & 75.8 & 81.9 & 76.0 & 82.2 & 76.4 & 0.0 & 0.0 & 90.4 & 86.7 \\ 
Latin & 49.2 & 33.6 & 53.9 & 36.2 & 51.3 & 33.3 & 54.0 & 35.5 & 53.9 & 35.4 & 0.0 & -0.8 & 67.2 & 54.5 \\ 
Slovene & 76.4 & 67.6 & 82.1 & 74.2 & 81.3 & 73.0 & 81.3 & 73.3 & 82.0 & 74.2 & -0.1 & 0.0 & 88.9 & 85.4 \\ \hline
Spanish\_ancora & 77.7 & 66.2 & 82.4 & 72.7 & 82.0 & 72.2 & 81.4 & 71.3 & 82.3 & 72.5 & -0.1 & -0.3 & 91.1 & 87.0 \\ 
Danish & 70.7 & 61.7 & 75.7 & 67.4 & 75.3 & 66.7 & 74.6 & 66.2 & 75.6 & 67.2 & -0.1 & -0.2 & 83.1 & 79.3 \\ 
Portuguese\_pud & 63.5 & 51.8 & 82.7 & 75.8 & 82.5 & 75.8 & 82.0 & 74.8 & 82.6 & 75.7 & -0.2 & -0.1 & 86.4 & 78.5 \\ 
Latin\_proiel & 59.2 & 46.2 & 61.5 & 47.4 & 60.9 & 47.1 & 60.2 & 46.0 & 61.3 & 47.2 & -0.2 & -0.2 & 80.9 & 75.4 \\ 
Slovak & 73.6 & 63.8 & 78.7 & 71.0 & 78.0 & 69.8 & 77.1 & 68.7 & 78.5 & 70.7 & -0.2 & -0.3 & 83.5 & 77.9 \\ 
\rowcolor{LightCyan} Hindi & 58.7 & 47.2 & 63.7 & 50.0 & 62.3 & 49.0 & 62.6 & 49.3 & 62.7 & 49.4 & -1.0 & -0.6 & 94.2 & 90.4 \\\hline
 Avg.  All   &  62.0  &  49.7  &  71.2  &  59.3  &  71.7  &  59.6  &  70.6  &  58.7  &  {\bf 72.1}  &  {\bf 60.0}  &  0.9 &	0.7  & 83.9  &  77.3  \\
\rowcolor{LightCyan}  Avg. Non-EU  &  46.6  &  32.0  &  57.1  &  40.9  &  60.1  &  43.1  &  57.8  &  41.9  &  {\bf 60.4}  &   {\bf 43.3 }   &  3.3 & 2.4 &  80.3  &  72.2  \\ \hline \hline
    \end{tabular}
    \caption{ Dependency parsing results, in terms of unlabeled attachment accuracy (UAS) and labeled attachment accuracy (LAS) after ignoring punctuations, on the Universal Dependencies v2 test sets~\protect\cite{universal_2} using  supervised part-of-speech tags. The results are sorted by their ``difference'' between the ensemble model and the baseline. The rows for non-European languages are highlighted with cyan. The rows that are highlighted by pink are the ones that the transfer model outperforms the supervised model. For all of the non-European datasets except ``hi'', our model outperforms significantly better in terms of UAS with $p < 0.001$ using McNemar's test. }
    \label{tab:reodering_main_results}
\end{table*}

\subsection{Results}
Table~\ref{tab:reodering_main_results} shows the results on the Universal Dependencies corpus~\cite{universal_2}. As shown in the table, the algorithm based on dominant dependency directions improves the accuracy on most of the non-European languages and performs slightly worse than the baseline model in the European languages. The ensemble model, in spite of its simplicity, improves over the baseline in most of the languages, leading to an average UAS improvement of $0.9$ for all languages and $3.3$ for non-European languages.  This improvement is very significant in many of the non-European languages; for example, from an LAS of $37.6$ to $52.7$ in Coptic, from a UAS of $44.9$ to $53.7$ in Basque, from a UAS of $40.6$ to $47.0$ in Chinese. Our model also outperforms the supervised models in Ukrainian and Latvian. That is an interesting indicator that for cases that the training data is very small for a language (37 sentences for Ukrainian, and 153 sentences for Latvian), our transfer approach outperforms the supervised model. 

\section{Analysis}
In this section, we briefly describe our analysis based on the results in the ensemble model and the baseline. For some languages such as Coptic, the number of dense projected dependencies is too small (two trees) such that the parser gives a worse learned model than a random baseline. For some other languages, such as Norwegian and Spanish, this number is too high (more than twenty thousand trees), such that the baseline model performs very well. 

The dominant dependency direction model generally performs better than the classifier. Our manual investigation shows that the classifier kept many of the dependency directions unchanged, while the dominant dependency direction model changed more directions. Therefore, the dominant direction model gives a higher recall with the expense of losing some precision. The training data for the reordering classifier is very noisy due to wrong alignments. We believe that the dominant direction model, besides its simplicity, is a more robust classifier for reordering, though the classifier is helpful in an ensemble setting. 

\begin{table}[!t]
   \centering
    \scriptsize
    \begin{tabular}{| l | c  | c || c | c  || c |  c | }
    \hline
 \multirow{2}{*}{data} & \multicolumn{2}{|c||}{ADJ} & \multicolumn{2}{c||}{NOUN} & \multicolumn{2}{c|}{VERB}\\ \cline{2-7}
  & {\tiny Base} & {\tiny Ens.} & {\tiny Base} & {\tiny Ens.} & {\tiny Base} & {\tiny Ens.}  \\\hline
 ar & \cellcolor{green!100} 40.4  &  \cellcolor{green!100} 46.7 & \cellcolor{green!95.0} 70.6  & \cellcolor{green!95.0} 72.5 & \cellcolor{green!100} 55.3  &  \cellcolor{green!100} 58.8\\ 
 ar\_pud & \cellcolor{green!100} 32.3  & \cellcolor{green!100} 39.7 & \cellcolor{green!100} 73.2  & \cellcolor{green!100}   75.9 & \cellcolor{green!100} 67.2  &   \cellcolor{green!100} 70.1\\ 
 bg & \cellcolor{green!25.0} 70.6  & \cellcolor{green!25.0} 71.1 & \cellcolor{green!20.0} 85.8  & \cellcolor{green!20.0} 86.2 & \cellcolor{green!15.0} 86.2  & \cellcolor{green!15.0}  86.5\\ 
 cop & \cellcolor{green!0.0} 0.0  &   \cellcolor{green!0.0} 0.0 & \cellcolor{green!100} 63.4  &  \cellcolor{green!100} 75.7 & \cellcolor{green!100} 64.6  &   \cellcolor{green!100} 76.4\\ 
 cs & \cellcolor{green!5.0} 64.8  & \cellcolor{green!5.0}  64.9 & \cellcolor{green!30.0} 77.9  & \cellcolor{green!30.0} 78.5 & \cellcolor{green!10.0} 76.5  &  \cellcolor{green!10.0} 76.7\\ 
 cs\_cac & \cellcolor{red!15.0} 66.0  &  \cellcolor{red!15.0} 65.7 & \cellcolor{green!40.0} 79.7  & \cellcolor{green!40.0} 80.5 & \cellcolor{green!15.0} 77.3  &  \cellcolor{green!15.0}  77.6\\ 
 cs\_cltt & \cellcolor{green!30.0} 55.9  &  \cellcolor{green!30.0} 56.5 & \cellcolor{green!40.0} 76.9  & \cellcolor{green!40.0}  77.7 & \cellcolor{green!30.0} 68.3  &\cellcolor{green!30.0}  68.9\\ 
 cs\_pud & \cellcolor{red!15.0} 71.2  &\cellcolor{red!15.0}   70.9 & \cellcolor{green!30.0} 79.4  & \cellcolor{green!30.0}  80 & \cellcolor{green!5.0} 80.2  & \cellcolor{green!5.0}  80.3\\ 
 da & \cellcolor{green!15.0} 70.9  &  \cellcolor{green!15.0} 71.2 & \cellcolor{red!10.0} 79.5  &\cellcolor{red!10.0}  79.3 & \cellcolor{green!5.0} 79.5  &  \cellcolor{green!5.0} 79.6\\ 
 de & \cellcolor{green!50.0} 65.7  &  \cellcolor{green!50.0} 66.7 & \cellcolor{green!10.0} 81.3  &\cellcolor{green!10.0}   81.5 & \cellcolor{green!25.0} 75.8  & \cellcolor{green!25.0} 76.3\\ 
 de\_pud & \cellcolor{green!55.0} 61.3  & \cellcolor{green!55.0}  62.4 & \cellcolor{green!0.0} 81.5  &\cellcolor{green!0.0}   81.5 & \cellcolor{green!10.0} 81.0  &  \cellcolor{green!10.0} 81.2\\ 
el & \cellcolor{green!25.0} 64.3  &   \cellcolor{green!25.0} 64.8 & \cellcolor{green!35.0} 79.8  &  \cellcolor{green!35.0}  80.5 & \cellcolor{green!10.0} 75.6  &  \cellcolor{green!10.0} 75.8\\ 
 en & \cellcolor{green!55.0} 77.7  & \cellcolor{green!55.0}  78.8 & \cellcolor{red!10.0} 70.6  & \cellcolor{red!10.0} 70.4 & \cellcolor{green!15.0} 81.0  &  \cellcolor{green!15.0} 81.3\\ 
 en\_lines & \cellcolor{green!15.0} 74.4  & \cellcolor{green!15.0} 74.7 & \cellcolor{green!10.0} 78.3  &  \cellcolor{green!10.0}  78.5 & \cellcolor{green!30.0} 82.2  &  \cellcolor{green!30.0} 82.8\\ 
 en\_partut & \cellcolor{green!10.0} 71.9  &  \cellcolor{green!10.0} 72.1 & \cellcolor{green!5.0} 76.6  &  \cellcolor{green!5.0} 76.7 & \cellcolor{green!5.0} 82.6  & \cellcolor{green!5.0}  82.7\\ 
 en\_pud & \cellcolor{green!55.0} 69.5  &   \cellcolor{green!55.0} 70.6 & \cellcolor{green!5.0} 75.4  & \cellcolor{green!5.0} 75.5 & \cellcolor{green!20.0} 81.2  & \cellcolor{green!20.0} 81.6\\ 
 es & \cellcolor{red!50.0} 75.6  & \cellcolor{red!50.0} 74.6 & \cellcolor{green!20.0} 88.0  & \cellcolor{green!20.0} 88.4 & \cellcolor{green!15.0} 80.6  & \cellcolor{green!15.0} 80.9\\ 
 es\_ancora & \cellcolor{green!5.0} 71.3  & \cellcolor{green!5.0}   71.4 & \cellcolor{green!0.0} 87.4  &  \cellcolor{green!0.0}  87.4 & \cellcolor{red!5.0} 83.0  &  \cellcolor{red!5.0} 82.9\\ 
 es\_pud & \cellcolor{red!10.0} 66.5  &  \cellcolor{red!10.0}66.3 & \cellcolor{green!5.0} 89.0  & \cellcolor{green!5.0} 89.1 & \cellcolor{green!0.0} 83.2  &   \cellcolor{green!0.0}83.2\\ 
 et & \cellcolor{green!5.0} 59.5  & \cellcolor{green!5.0} 59.6 & \cellcolor{red!5.0} 59.6  & \cellcolor{red!5.0} 59.5 & \cellcolor{green!5.0} 75.4  &  \cellcolor{green!5.0} 75.5\\ 
 eu & \cellcolor{green!100} 31.1  & \cellcolor{green!100} 35.4 & \cellcolor{green!100} 37.6  & \cellcolor{green!100} 47.9 & \cellcolor{green!100} 52.4  & \cellcolor{green!100}  61.2\\ 
 fa & \cellcolor{green!100} 46.2  & \cellcolor{green!100}   51.6 & \cellcolor{green!100.0} 68.7  &  \cellcolor{green!100.0} 70.7 & \cellcolor{green!100} 53.7  &\cellcolor{green!100}   59.7\\ 
 fi & \cellcolor{green!25.0} 65.8  &   \cellcolor{green!25.0} 66.3 & \cellcolor{green!35.0} 61.8  & \cellcolor{green!35.0} 62.5 & \cellcolor{green!0.0} 70.5  & \cellcolor{green!0.0} 70.5\\ 
 fi\_ftb & \cellcolor{green!40.0} 64.7  & \cellcolor{green!40.0} 65.5 & \cellcolor{green!20.0} 64.7  &  \cellcolor{green!20.0} 65.1 & \cellcolor{green!15.0} 69.2  & \cellcolor{green!15.0} 69.5\\ 
 fi\_pud & \cellcolor{green!65.0} 58.1  & \cellcolor{green!65.0} 59.4 & \cellcolor{green!15.0} 63.8  &  \cellcolor{green!15.0} 64.1 & \cellcolor{green!10.0} 74.6  &  \cellcolor{green!10.0} 74.8\\ 
 fr & \cellcolor{green!25.0} 74.1  &  \cellcolor{green!25.0} 74.6 & \cellcolor{green!10.0} 87.3  & \cellcolor{green!10.0} 87.5 & \cellcolor{green!40.0} 81.9  & \cellcolor{green!40.0} 82.7\\ 
 fr\_partut & \cellcolor{green!35.0} 72.2  & \cellcolor{green!35.0} 72.9 & \cellcolor{green!20.0} 88.4  & \cellcolor{green!20.0} 88.8 & \cellcolor{green!35.0} 83.1  & \cellcolor{green!35.0}  83.8\\ 
fr\_pud & \cellcolor{red!10.0} 71.3  &  \cellcolor{red!10.0} 71.1 & \cellcolor{green!5.0} 88.7  &   \cellcolor{green!5.0} 88.8 & \cellcolor{green!5.0} 81.0  & \cellcolor{green!5.0} 81.1\\ 
 fr\_sequoia & \cellcolor{green!0.0} 72.0  &  \cellcolor{green!0.0} 72.0 & \cellcolor{green!5.0} 86.5  &\cellcolor{green!5.0}   86.6 & \cellcolor{red!10.0} 82.2  &  \cellcolor{red!10.0} 82.0\\ 
 he & \cellcolor{green!100} 64.7  & \cellcolor{green!100}  69.1 & \cellcolor{green!100} 75.6  &  \cellcolor{green!100}  77.8 & \cellcolor{green!100} 68.1  & \cellcolor{green!100} 70.6\\ 
 hi & \cellcolor{green!60.0} 22.3  &  \cellcolor{green!60.0} 23.5 & \cellcolor{red!50.0} 75.9  &\cellcolor{red!50.0}   74.9 & \cellcolor{green!20.0} 57.5  & \cellcolor{green!20.0} 57.9\\ 
 hi\_pud & \cellcolor{green!60.0} 48.1  & \cellcolor{green!60.0} 49.3 & \cellcolor{green!5.0} 67.8  &   \cellcolor{green!5.0} 67.9 & \cellcolor{green!100} 56.6  & \cellcolor{green!100} 58.7\\ 
hr & \cellcolor{red!25.0} 72.3  & \cellcolor{red!25.0}  71.8 & \cellcolor{green!10.0} 82.2  &  \cellcolor{green!10.0}  82.4 & \cellcolor{green!35.0} 83.1  &  \cellcolor{green!35.0} 83.8\\ 
 hu & \cellcolor{green!40.0} 42.5  & \cellcolor{green!40.0}   43.3 & \cellcolor{green!35.0} 71.8  & \cellcolor{green!35.0}  72.5 & \cellcolor{green!5.0} 73.6  & \cellcolor{green!5.0} 73.7\\ 
id & \cellcolor{green!100} 63.2  & \cellcolor{green!100} 67.3 & \cellcolor{green!100} 70.7  &  \cellcolor{green!100} 74.5 & \cellcolor{green!85.0} 78.0  & \cellcolor{green!85.0} 79.7\\ 
 it & \cellcolor{green!95.0} 61.4  &   \cellcolor{green!95.0} 63.3 & \cellcolor{green!0.0} 89.1  & \cellcolor{green!0.0}  89.1 & \cellcolor{green!10.0} 85.2  & \cellcolor{green!10.0} 85.4\\ 
 it\_pud & \cellcolor{green!15.0} 71.7  & \cellcolor{green!15.0} 72.0 & \cellcolor{green!0.0} 90.7  & \cellcolor{green!0.0} 90.7 & \cellcolor{green!5.0} 87.1  & \cellcolor{green!5.0}  87.2\\ 
 ja & \cellcolor{green!100} 52.8  & \cellcolor{green!100}  59.5 & \cellcolor{green!75.0} 73.1  & \cellcolor{green!75.0} 74.6 & \cellcolor{green!70.0} 65.1  &  \cellcolor{green!70.0} 66.5\\ 
ja\_pud & \cellcolor{green!100} 60.4  & \cellcolor{green!100}   65.4 & \cellcolor{green!55.0} 71.5  & \cellcolor{green!55.0} 72.6 & \cellcolor{green!80.0} 66.7  & \cellcolor{green!80.0} 68.3\\ 
 ko & \cellcolor{red!100} 55.7  &  \cellcolor{red!100}  52.9 & \cellcolor{green!40.0} 23.5  &  \cellcolor{green!40.0}  24.3 & \cellcolor{green!95.0} 52.4  & \cellcolor{green!95.0} 54.3\\ 
la & \cellcolor{green!25.0} 35.1  &  \cellcolor{green!25.0} 35.6 & \cellcolor{green!30.0} 43.8  & \cellcolor{green!30.0} 44.4 & \cellcolor{red!15.0} 58.8  & \cellcolor{red!15.0}  58.5\\ 
 la\_ittb & \cellcolor{red!25.0} 57.9  & \cellcolor{red!25.0} 57.4 & \cellcolor{green!50.0} 65.5  & \cellcolor{green!50.0} 66.5 & \cellcolor{green!5.0} 63.5  & \cellcolor{green!5.0} 63.6\\ 
la\_proiel & \cellcolor{green!10.0} 55.2  &  \cellcolor{green!10.0} 55.4 & \cellcolor{red!10.0} 61.8  &  \cellcolor{red!10.0} 61.6 & \cellcolor{red!10.0} 64.3  & \cellcolor{red!10.0} 64.1\\ 
lt & \cellcolor{green!100} 54.0  &  \cellcolor{green!100} 57.1 & \cellcolor{green!70.0} 70.8  &  \cellcolor{green!70.0} 72.2 & \cellcolor{green!0.0} 69.7  &\cellcolor{green!0.0}   69.7\\ 
lv & \cellcolor{green!75.0} 58.7  & \cellcolor{green!75.0} 60.2 & \cellcolor{green!10.0} 57.0  &  \cellcolor{green!10.0} 57.2 & \cellcolor{green!15.0} 70.3  & \cellcolor{green!15.0} 70.6\\ 
nl & \cellcolor{green!100} 57.7  & \cellcolor{green!100}  61.3 & \cellcolor{red!10.0} 81.9  & \cellcolor{red!10.0} 81.7 & \cellcolor{green!40.0} 66.4  & \cellcolor{green!40.0}   67.2\\ 
 nl\_lassysmall & \cellcolor{green!60.0} 46.4  & \cellcolor{green!60.0}   47.6 & \cellcolor{green!10.0} 79.8  & \cellcolor{green!10.0}  80.0& \cellcolor{red!5.0} 75.4  & \cellcolor{red!5.0} 75.3\\ 
 no\_bokmaal & \cellcolor{red!5.0} 76.0  &   \cellcolor{red!5.0} 75.9 & \cellcolor{green!0.0} 83.4  & \cellcolor{green!0.0} 83.4 & \cellcolor{green!10.0} 84.2  & \cellcolor{green!10.0} 84.4\\ 
 no\_nynorsk & \cellcolor{green!40.0} 69.7  &   \cellcolor{green!40.0} 70.5 & \cellcolor{green!20.0} 81.4  &   \cellcolor{green!20.0} 81.8 & \cellcolor{green!15.0} 79.6  & \cellcolor{green!15.0}  79.9\\ 
 pl & \cellcolor{green!55.0} 66.1  &  \cellcolor{green!55.0} 67.2 & \cellcolor{green!10.0} 79.2  & \cellcolor{green!10.0}   79.4 & \cellcolor{green!0.0} 85.2  & \cellcolor{green!0.0}   85.2\\ 
 pt & \cellcolor{green!60.0} 72.1  & \cellcolor{green!60.0} 73.3 & \cellcolor{green!5.0} 88.7  &  \cellcolor{green!5.0} 88.8 & \cellcolor{red!10.0} 82.9  &   \cellcolor{red!10.0} 82.7\\ 
 pt\_br & \cellcolor{green!0.0} 39.5  &  \cellcolor{green!0.0} 39.5 & \cellcolor{green!0.0} 88.2  & \cellcolor{green!0.0} 88.2 & \cellcolor{red!5.0} 77.8  &  \cellcolor{red!5.0}77.7\\ 
 pt\_pud & \cellcolor{red!45.0} 61.4  & \cellcolor{red!45.0} 60.5 & \cellcolor{red!10.0} 89.0  & \cellcolor{red!10.0}  88.8 & \cellcolor{green!0.0} 81.2  &\cellcolor{green!0.0}   81.2\\ 
 ro & \cellcolor{green!40.0} 55.6  & \cellcolor{green!40.0}  56.4 & \cellcolor{green!10.0} 79.3  &  \cellcolor{green!10.0}  79.5 & \cellcolor{green!0.0} 80.3  & \cellcolor{green!0.0} 80.3\\ 
 ru & \cellcolor{green!40.0} 52.3  &  \cellcolor{green!40.0} 53.1 & \cellcolor{green!30.0} 77.9  &  \cellcolor{green!30.0} 78.5 & \cellcolor{green!10.0} 79.8  & \cellcolor{green!10.0}  80.0\\ 
 ru\_pud & \cellcolor{green!5.0} 64.4  &  \cellcolor{green!5.0} 64.5 & \cellcolor{green!30.0} 83.1  &\cellcolor{green!30.0}  83.7 & \cellcolor{green!25.0} 81.9  &   \cellcolor{green!25.0} 82.4\\ 
 ru\_syntagrus & \cellcolor{red!20.0} 57.3  &  \cellcolor{red!20.0} 56.9 & \cellcolor{green!25.0} 78.7  &  \cellcolor{green!25.0} 79.2 & \cellcolor{green!15.0} 74.3  &  \cellcolor{green!15.0} 74.6\\ 
sk & \cellcolor{green!0.0} 69.5  & \cellcolor{green!0.0} 69.5 & \cellcolor{red!10.0} 80.5  &   \cellcolor{red!10.0}80.3 & \cellcolor{red!25.0} 84.0  & \cellcolor{red!25.0}  83.5\\ 
 sl & \cellcolor{red!50.0} 73.6  &   \cellcolor{red!50.0} 72.6 & \cellcolor{green!5.0} 83.3  &  \cellcolor{green!5.0} 83.4 & \cellcolor{green!0.0} 85.2  &  \cellcolor{green!0.0}85.2\\ 
sl\_sst & \cellcolor{green!50.0} 60.6  &   \cellcolor{green!50.0} 61.6 & \cellcolor{green!10.0} 69.4  &\cellcolor{green!10.0}  69.6 & \cellcolor{red!15.0} 67.4  & \cellcolor{red!15.0} 67.1\\ 
 sv & \cellcolor{red!5.0} 77.0  & \cellcolor{red!5.0} 76.9 & \cellcolor{green!5.0} 82.8  &  \cellcolor{green!5.0}  82.9 & \cellcolor{green!15.0} 81.1  &   \cellcolor{green!15.0} 81.4\\ 
 sv\_lines & \cellcolor{green!10.0} 78.7  &   \cellcolor{green!10.0} 78.9 & \cellcolor{green!0.0} 85.1  &  \cellcolor{green!0.0} 85.1 & \cellcolor{green!10.0} 83.1  &  \cellcolor{green!10.0} 83.3\\ 
 sv\_pud & \cellcolor{green!5.0} 77.2  & \cellcolor{green!5.0}  77.3 & \cellcolor{green!0.0} 83.4  &  \cellcolor{green!0.0} 83.4 & \cellcolor{green!15.0} 83.8  &\cellcolor{green!15.0}   84.1\\ 
 tr & \cellcolor{green!100} 42.3  & \cellcolor{green!100}  46.8 & \cellcolor{red!75.0} 49.4  & \cellcolor{red!75.0} 47.9 & \cellcolor{green!100} 48.0  & \cellcolor{green!100}  51.5\\ 
 tr\_pud & \cellcolor{green!100} 43.2  &  \cellcolor{green!100} 46.7 & \cellcolor{green!100.0} 50.0  & \cellcolor{green!100.0}  52.0 & \cellcolor{green!100} 49.5  &  \cellcolor{green!100} 53.5\\ 
uk & \cellcolor{red!25.0} 49.3  & \cellcolor{red!25.0}  48.8 & \cellcolor{green!25.0} 64.1  &\cellcolor{green!25.0}  64.6 & \cellcolor{green!20.0} 71.9  &  \cellcolor{green!20.0} 72.3\\ 
 vi & \cellcolor{green!100} 31.5  & \cellcolor{green!100} 35.7 & \cellcolor{green!100} 50.6  & \cellcolor{green!100}  56.5 & \cellcolor{green!100} 55.1  & \cellcolor{green!100} 58.3\\ 
 zh & \cellcolor{green!100} 47.7  & \cellcolor{green!100} 52.1 & \cellcolor{green!100} 47.5  &  \cellcolor{green!100} 56.4 & \cellcolor{green!100} 43.0  & \cellcolor{green!100}  45.7\\ \hline
    \end{tabular}
    \caption{Unlabeled attachment f-score of POS tags as heads for the baseline and the reordering ensemble model. We show the green color for improvement, and the red color for worse result in the ensemble model; the darkness of the color indicates the level of difference.}
    \label{tab:head_pos_analysis}
\end{table}

Our detailed analysis show that we are able to improve the head dependency relation for the three most important head POS tags in the dependency grammar. We see that this improvement is more consistent for all non-European languages. Table~\ref{tab:head_pos_analysis} shows the differences in parsing f-score of dependency relations for adjectives, nouns and verbs as the head. As we see in the Table, we are able to improve the head dependency relation for the three most important head POS tags in the dependency grammar. We see that this improvement is more consistent for all non-European languages. We skip the details of those analysis due to space limitations. More thorough analysis can be found in ~\cite[Chapter~6]{rasooli2019cross}.

 For a few number of languages such as Vietnamese, the best model, even though improves over a strong baseline, still lacks enough accuracy to be considered as a reliable parser in place of a supervised model. We believe that more research on those language will address the mentioned problem. Our current model relies on supervised part-of-speech tags. Future work should study using transferred part-of-speech tags instead of supervised tags, leading to a much more realistic scenario for low-resource languages. 
 
 We have also calculated the POS trigram cosine similarity between the target language gold standard treeebanks, and the three source training datasets (original, and the two reordered datasets). In all of the non-European languages, the cosine similarity of the reordered datasets improved with different values in the range of $(0.002,0.02)$. For Czech, Portuguese, German, Greek, English, Romanian, Russian, and Slovak, both of the reordered datasets slightly decreased the trigram cosine similarity. For other languages, the cosine similarity was roughly the same.

 \section{Conclusion}
We have described a cross-lingual dependency transfer method that takes into account the problem of word order differences between the source and target languages. We have shown that applying projection-driven reordering improves the accuracy of non-European languages while maintaining the high accuracies in European languages. The focus of this paper is primarily of dependency parsing. Future work should investigate the effect of our proposed reordering methods on truly low-resource machine translation.
 
 \section*{Acknowledgements}
We deeply thank the anonymous reviewers for their useful feedback and comments. 
\bibliography{refs}
\bibliographystyle{acl_natbib}
\end{document}